\documentclass[10pt,journal,compsoc]{IEEEtran}
\usepackage{amsmath,amsfonts}
\usepackage{algorithmic}
\usepackage{algorithm}
\usepackage{array}
\usepackage[caption=false,font=normalsize,labelfont=sf,textfont=sf]{subfig}
\usepackage{textcomp}
\usepackage{stfloats}
\usepackage{url}
\usepackage{verbatim}
\usepackage{graphicx}
\usepackage{cite}
\usepackage{hyperref}
\usepackage{cleveref}
\usepackage{booktabs}
\usepackage{multirow}
\usepackage{mdwlist}
\usepackage{float}
\usepackage{subfig}
\usepackage{color,xcolor}
\usepackage{colortbl}
\usepackage{fancyhdr}
\usepackage{marvosym}
\usepackage{amssymb}
\usepackage{ragged2e}
\hypersetup{
hidelinks,
colorlinks=true,
linkcolor=red,
citecolor=blue
}

\definecolor{best}{RGB}{242, 158, 156}
\definecolor{second}{RGB}{248, 206, 160}
\definecolor{third}{RGB}{255, 254, 166}

\begin{document}

\title{Laser: Efficient Language-Guided Segmentation in Neural Radiance Fields}

\author{Xingyu Miao, Haoran~Duan*\Letter~\IEEEmembership{Member,~IEEE}, Yang Bai,~Tejal~Shah,~Jun~Song, \\ ~Yang~Long\Letter,~\IEEEmembership{Senior Member,~IEEE}, ~Rajiv~Ranjan,~\IEEEmembership{Fellow,~IEEE}, and~Ling~Shao,~\IEEEmembership{Fellow,~IEEE}

\IEEEcompsocitemizethanks{
\IEEEcompsocthanksitem Xingyu Miao and Yang Long are with the Department of Computer Science, Durham University, UK. E-mail: xingyu.miao@durham.ac.uk; yang.long@ieee.org.
\IEEEcompsocthanksitem Yang Bai is with the Institute of High Performance Computing (IHPC), ASTAR, Singapore. E-mail: bai\_yang@ihpc.a-star.edu.sg
\IEEEcompsocthanksitem Jun Song is with School of Computer Science, China University of Geosciences, Wuhan, 430074, P. R. China. E-mail: songjun@cug.edu.cn
\IEEEcompsocthanksitem Rajiv Ranjan, Haoran Duan and Tejal Shah are with the School of Computing, Newcastle University, UK. E-mail: rranjans@gmail.com, haoran.duan@ieee.org, tejal.shah@newcastle.ac.uk
\IEEEcompsocthanksitem Ling Shao is with the UCAS-Terminus AI Lab, University of Chinese Academy of Sciences, Beijing 100049, China. E-mail:ling.shao@ieee.org. 
\IEEEcompsocthanksitem Yang Long and Haoran Duan are the corresponding author.
\IEEEcompsocthanksitem Xingyu Miao* and Haoran Duan* have equal contribution.
}
\thanks{Manuscript submitted March 15, 2024;}}
% \IEEEcompsocthanksitem Ling Shao is with the UCAS-Terminus AI Lab, University of Chinese Academy of Sciences, Beijing 100049, China. E-mail:ling.shao@ieee.org. 

\markboth{IEEE TRANSACTIONS ON PATTERN ANALYSIS AND MACHINE INTELLIGENCE}
{Shell \MakeLowercase{\textit{et al.}}: Bare Demo of IEEEtran.cls for Computer Society Journals}


\IEEEtitleabstractindextext{%
\begin{abstract}
\justifying
In this work, we propose a method that leverages CLIP feature distillation, achieving efficient 3D segmentation through language guidance. Unlike previous methods that rely on multi-scale CLIP features and are limited by processing speed and storage requirements, our approach aims to streamline the workflow by directly and effectively distilling dense CLIP features, thereby achieving precise segmentation of 3D scenes using text. To achieve this, we introduce an adapter module and mitigate the noise issue in the dense CLIP feature distillation process through a self-cross-training strategy. Moreover, to enhance the accuracy of segmentation edges, this work presents a low-rank transient query attention mechanism. To ensure the consistency of segmentation for similar colors under different viewpoints, we convert the segmentation task into a classification task through label volume, which significantly improves the consistency of segmentation in color-similar areas. We also propose a simplified text augmentation strategy to alleviate the issue of ambiguity in the correspondence between CLIP features and text. Extensive experimental results show that our method surpasses current state-of-the-art technologies in both training speed and performance. Our code is available on: \url{https://github.com/xingy038/Laser.git}.
\end{abstract}

}

\maketitle

\section{Introduction}
\IEEEPARstart{A}{dvancements} in Neural Radiance Fields (NeRFs) \cite{mildenhall2021nerf} and 3D Gaussian splatting (3D GS)\cite{kerbl3Dgaussians} have driven progress in novel view synthesis of 3D scenes in recent years. However, these models often focus primarily on geometry and appearance modeling with limited consideration of semantic information, thereby affecting the overall understanding of 3D scenes. Comprehensive understanding of 3D scenes is crucial for applications such as robot navigation, or autonomous driving. Several endeavors \cite{siddiqui2023panoptic, bhalgat2023contrastive} have aimed to enrich 3D scenes with multi-view semantic information. Despite these advancements, the scarcity of adequately annotated 3D scene datasets continues to pose a significant hurdle. An effective approach to address this issue involves adopting open vocabulary segmentation, which facilitates the identification and classification of diverse semantic regions and objects within these scenes. However, this strategy introduces its own set of substantial challenges, chiefly the requirement for an in-depth understanding of both natural language processing and the accurate representation of objects in 3D environments.

\begin{figure}
    \centering
    \includegraphics[width=\linewidth]{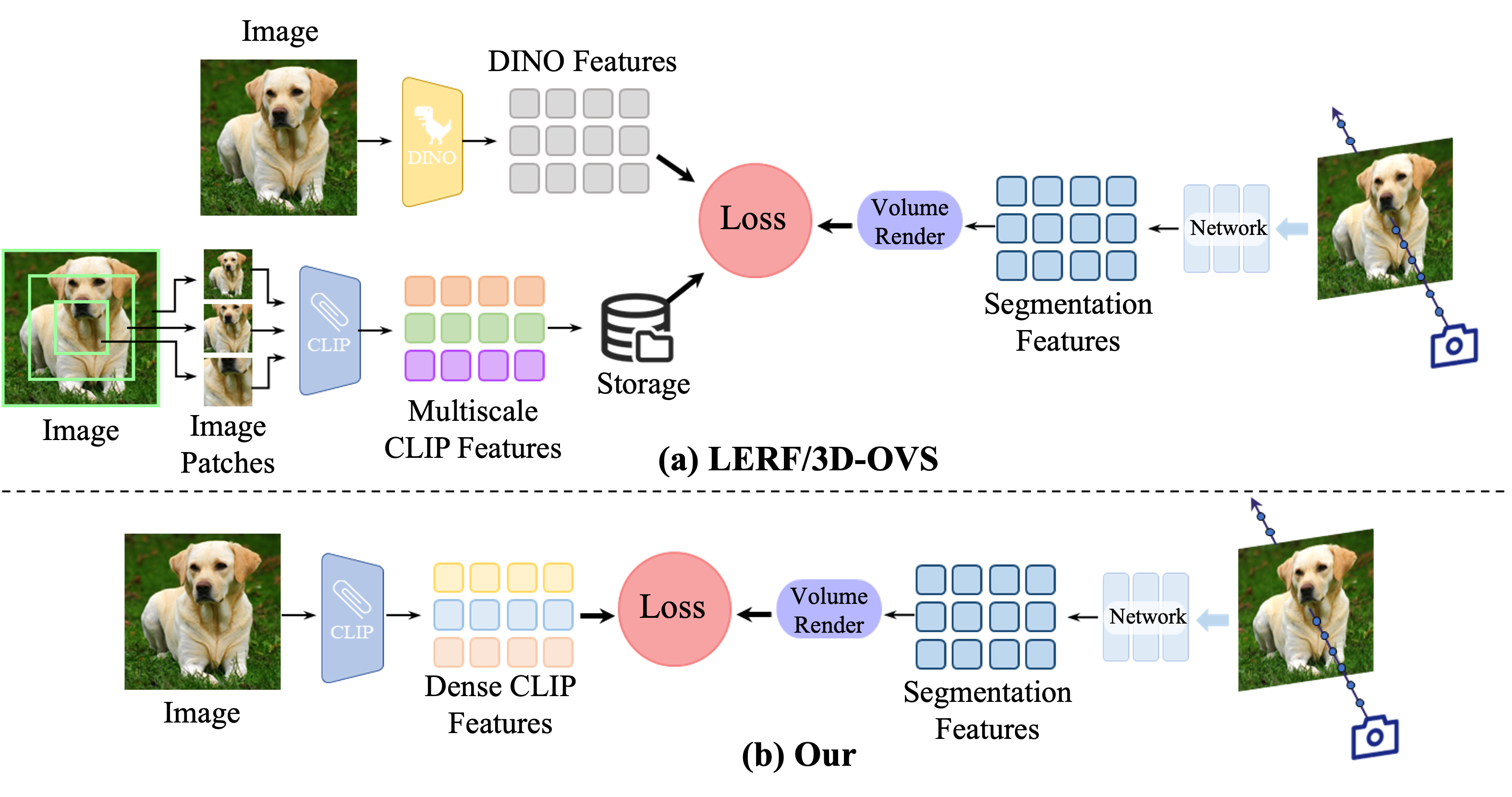}
    \caption{\textbf{Workflows} of existing methods and Ours: (a) The core of the LERF/3D-OVS process initially adopts a cutting strategy that subdivides the training images into patches of different sizes. These patches are then fed into the CLIP encoder to extract multi-scale CLIP features, which are subsequently saved. At the same time, the original image is also input into DINO to extract DINO features. Afterwards, these multi-scale CLIP features and DINO features participate together in the segmentation branch optimization process of NeRF. Such a process demands significantly high computational and storage resources. (b) Compared to previous methods, our process only requires the input of training images into a modified CLIP encoder, after which it can predict dense CLIP features. Utilizing these features, the segmentation branch of NeRF can be optimized efficiently, while significantly reducing the consumption of computational and storage resources.}
    \label{fig:teaser}
\end{figure}

Recent work \cite{lerf2023,liu2023weakly,kobayashi2022decomposing} distills features from existing visual language models, such as CLIP \cite{radford2021learning}, into 3D scenes for performing open-vocabulary query tasks. This task aims to enable human users to interact with large language models (LLM) \cite{li2023blip,li2022blip,devlin2018bert,radford2021learning} for scene editing intuitively. In comparison to traditional semantic labeling methods, the language features of visual language models provide a more comprehensive semantic understanding, especially encompassing objects in the long-tail distribution, making them more applicable in practical scenarios. However, accurately integrating language embeddings into the current 3D scene representation while maintaining efficiency and visual quality poses a few significant challenges. One challenge is aligning the learned 3D points and CLIP embeddings during rendering since CLIP embeddings are represented at the image level rather than the pixel level. To address this issue, LERF \cite{lerf2023} and 3D-OVS \cite{liu2023weakly} employ a patch cropping approach to obtain CLIP embeddings at different scales. However, this approach introduces a challenge related to clarity or precision in identifying semantic scales associated with the same 3D position. For instance, if we consider a point on the left fan of a graphics card, different levels of specificity in textual queries can lead to the identification of regions that correspond to varying scales of detail. Specifically, text queries such as "graphics card," "left side of the graphics card," and "fan on the left side of the graphics card" may each highlight areas that are relevant at three distinct levels of granularity. 
To address this problem, the LERF simply averages the multi-scale features, while 3D-OVS uses a 3D selection volume to choose the most appropriate scale. Despite their innovative approaches, both methods significantly hamper processing efficiency. This inefficiency is evident in the pre-processing phase, where prioritizing the extraction and storage of multi-scale features necessitates a substantial storage capacity—approximately 35GB for the features of each scene.Furthermore, multi-scale patch features bring another issue where they cannot fully include a segmented object, either including other objects in one patch or the objects in the patch are incomplete. 
Due to these inaccurate CLIP features, the 3D language embedding fields lack clear boundaries and contain a significant amount of noise. To alleviate this issue, these methods extract pre-trained DINO features \cite{caron2021emerging} concurrently during training and align them with 3D segmentation features. However, this approach leads to a further increase in training time and extra information requirements (The workflow as shown in \Cref{fig:teaser} (a)).

Driven by the goal of achieving rapid training speeds without compromising on accuracy in 3D language embedding fields, this paper presents Laser - an efficient end-to-end framework for segmenting 3D neural radiance fields, designed to balance high performance with computational efficiency (The overall workflows of our work as shown in \Cref{fig:teaser} (b)). In order to obtain pixel-level features from CLIP, we draw inspiration from recent advancements in CLIP dense prediction research, such as MaskCLIP \cite{ding2023maskclip}, CAT-Seg \cite{cho2023catseg}, and SCLIP \cite{wang2023sclip}. These methods modify the final pooling layer within the CLIP encoder to derive dense image embeddings. However, these methods introduce significant noise without a mechanism for refinement, potentially resulting in inaccurate predictions. Although aligning these dense features with 3D segmentation features directly can yield a roughly segmented image, there are still significant amounts of noise present. To address this issue, we propose a novel adapter that effectively reduces the noise present in the segmented images through the incorporation of a loss term and a self-cross-training strategy during training. Although this can alleviate the noise problem and obtain relatively accurate segmentation maps, it cannot guarantee the consistency of different views in the same scene. To maintain segmentation consistency across different views, we introduce label volumes to transform the segmentation problem into a classification task. In this way, we can classify rays within the segmentation volume, thus ensuring consistent segmentation results under different views. However, we observe that two query texts in the same category area may become ambiguous due to variations in lighting and image color. This ambiguity may lead to noise in the same segmentation region (i.e. regions segmented as the same category may include other categories). To alleviate this issue, we propose a novel low-rank transient query attention and a simplified text augmentation strategy to enhance features. In summary, the contributions of our work are as follows:

\begin{itemize}
  \item We review prior 3D language-guided segmentation methods that primarily target broad modality alignment. To attain finer alignment, we introduce a streamlined framework designed for efficient segmentation through 3D language guidance.
  
  \item We introduce a novel adapter module and self-crossing method to reduce the noise impact caused by dense CLIP features.

  \item We introduce a novel low-rank transient query attention model that significantly reduces the computational complexity required by vanilla attention mechanisms when processing 3D features.

  \item We carefully propose a simplified text enhancement strategy that significantly reduces the ambiguity of similarity calculation between CLIP features and text features, thereby effectively improving the accuracy of feature matching.
\end{itemize}

\section{Related work}

\subsection{Neural Rendering}
Image-based novel view synthesis methods typically employ input images to integrate pixel information, and their performance can be enhanced through multi-view approaches. Light field and Lomography rendering utilize filtering and interpolating ray samples to generate novel views without the need for an explicit geometric model. However, they necessitate capturing and storing numerous views for high-quality rendering \cite{chai2000plenoptic,kalantari2016learning,gortler1996lumigraph,levoy1996light,wu2017light}. Many approaches leverage explicit proxy geometry to achieve high-quality rendering with only a few input images \cite{buehler2001unstructured,debevec1996modeling,flynn2016deepstereo,hedman2018deep,hedman2016scalable,kalantari2016learning,kopf2014first,penner2017soft,riegler2020free,riegler2021stable}. However, these methods may suffer from untextured regions, highlights, reflections, and repetitive patterns, posing challenges for accurately estimating scene geometry. To address these issues, previous works have explored techniques such as local warps, soft 3D reconstruction, and learning-based approaches.

Recently, neural implicit representation methods like NeRF \cite{liu2020neural,mildenhall2022nerf,xian2021space,mildenhall2021nerf,xiangli2022bungeenerf,xu2022point,yu2021pixelnerf} have demonstrated significant potential for high-quality rendering. NeRF employs multi-layer perceptrons (MLPs) to implicitly represent continuous scenes, yielding impressive view synthesis results. The advent of NeRF marks substantial progress in the neural network representation of 3D scenes. However, these methods require training a separate model for each scene, and optimizing these models entails varying training times. Several recent methods have effectively improved rendering and training speed through explicit and hybrid scene representation \cite{chen2022tensorf, muller2022instant, sun2022direct, yu2021plenoctrees}. These methods utilize explicit spatial data structures to store explicit scene data \cite{fridovich2022plenoxels, yu2021plenoctrees} or features decoded by micro MLP \cite{chen2022tensorf, muller2022instant, sun2022direct}. This separates the model's capacity from its speed, allowing real-time rendering of high-quality images \cite{muller2022instant}. However, embedding language representations directly onto the neural field of implicit expressions is not an easy task, as a large number of sampling points require extensive memory usage, resulting in a sharp decline in training and rendering performance.

\subsection{2D Language Segmentation}
The goal of 2D language segmentation is to understand arbitrary categories in images with textual descriptions. In the previous works, ZS3Net \cite{bucher2019zero} is a pioneering work that utilizes word embeddings of unseen classes through a generative model to generate pixel-level features. SPNet \cite{xian2019semantic} uses word embeddings (e.g., word2vec \cite{mikolov2013efficient}) to achieve alignment of semantic and visual features. GroupViT \cite{xu2022groupvit} proceeds by grouping segmentation masks directly from text supervision. In the recent methods, they utilize pre-trained CLIP \cite{radford2021learning} for open-lexical semantic segmentation. LSeg \cite{li2022languagedriven} adopts a strategy of aligning pixel embeddings with text embeddings of corresponding semantic classes generated by the CLIP text encoder. Different from pixel-level LSeg, OpenSeg \cite{ghiasi2022scaling} proposes a method to achieve segment-level visual feature alignment with text embeddings through a regional word basis. 3D-OVS \cite{liu2023weakly} is the first work on implicit neural field open vocabulary segmentation. However, this method is essentially trained and tested on a closed vocabulary set. It achieves effective segmentation of 3D scenes by aligning multi-scale CLIP features with 3D segmentation features. On the contrary, to improve efficiency, we adopt the latest CLIP dense prediction method to obtain dense CLIP features \cite{ding2023maskclip}, avoiding using memory-intensive and computationally slow multi-scale CLIP features.

\subsection{3D Language Fields}
Due to the challenges in accurately identifying 3D regions, incorporating specific semantics into the implicit representation of NeRF based on MLP is extremely difficult. Some methods \cite{fu2022panoptic, bhalgat2023contrastive, shen2023anything, xu2023open} add new network branches and attempt to integrate semantic information into NeRF. For instance, Semantic NeRF \cite{zhi2021place} jointly encodes semantics with appearance and geometric shapes in NeRF to achieve novel semantic view synthesis. Neural Feature Fusion Fields \cite{tschernezki2022neural} use self-supervised 2D semantic features to distill 3D features. Recent work explores integrating language into 3D scenes for segmentation. For example, DFF \cite{kobayashi2022decomposing} adds a branch for language prediction, attempting to segment unseen text labels during training by fitting the feature map of LSeg \cite{li2022languagedriven} into NeRF. LERF \cite{lerf2023} is the first method to embed CLIP features \cite{radford2021learning} into NeRF, utilizing powerful CLIP representations for text-based 3D queries. Additionally, they use DINO features \cite{caron2021emerging} to supervise LERF and enhance its performance. 3D-OVS \cite{liu2023weakly} optimizes LERF and achieves open-vocabulary segmentation functionality. However, the low semantic embedding quality of these methods requires the introduction of additional semantic information such as DINO features \cite{caron2021emerging} or SAM features \cite{kirillov2023segment}, and the time cost during training is high. In addition, with the rise of 3D GS explicit representation, some recent segmentation methods based on 3D GS have also shown excellent results, such as Langsplat \cite{qin2024langsplat}, LEGaussians \cite{shi2024language}, and Feature 3dgs \cite{zhou2024feature}. However, these methods still use the data processing method of 3D-OVS. Although the real-time rendering advantage of 3D GS improves the training speed, there is still a lot of redundancy in the preprocessing stage. In contrast, our proposed Laser can provide an efficient 3D language scene representation.

\begin{figure}
    \centering
    \includegraphics[width=0.8\linewidth]{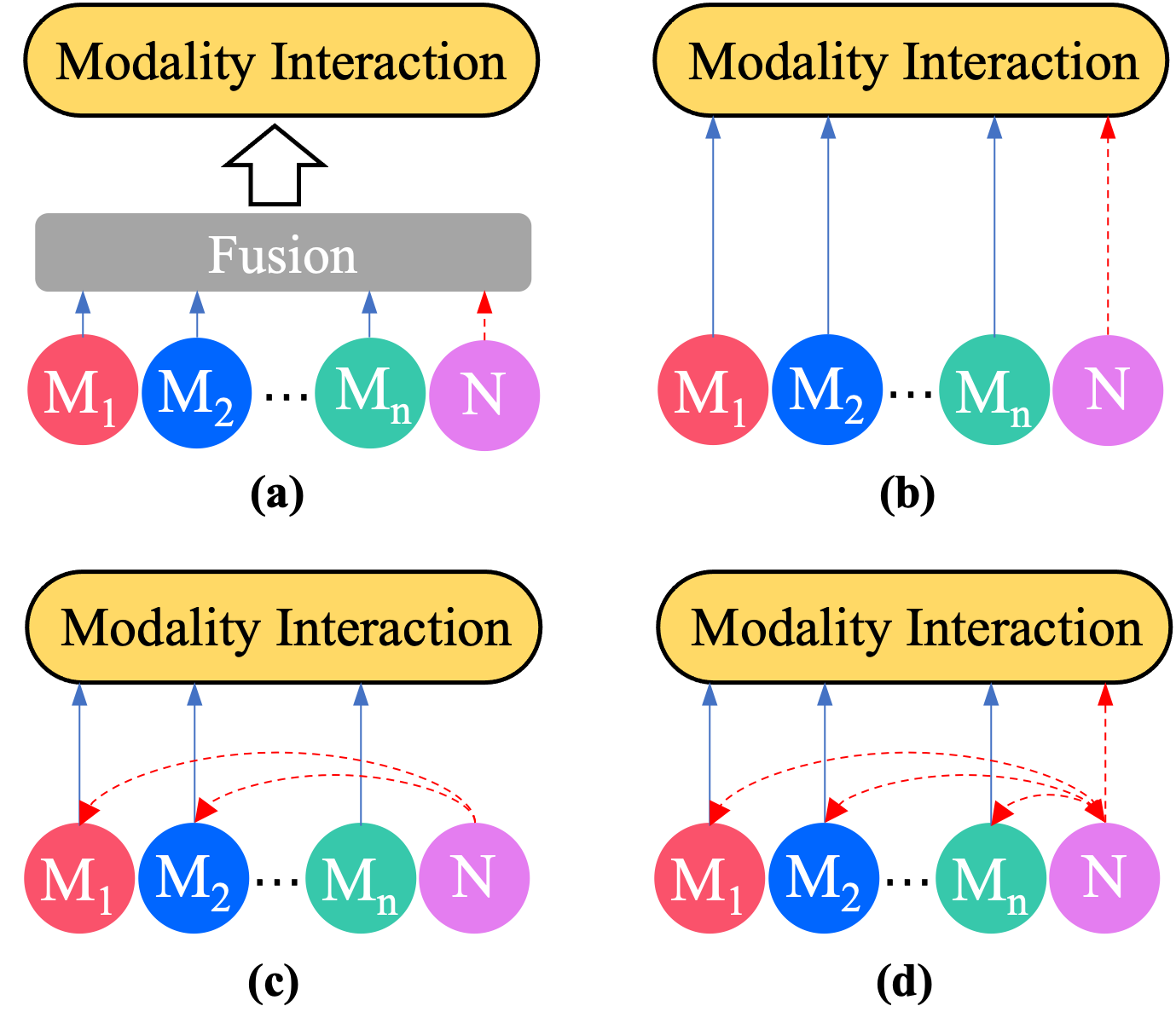}
    \caption{\textbf{Interaction comparison of different modality.} M=modality, N=new modality. (a) Fusion of features from different modalities and then interactive processing. When new modalities are included, retraining is required. (b) Directly interact with features of different modalities. When new modalities are added, they also need to be retrained. (c) Directly interact with the features of different modalities. After introducing a new modality, similar modal features can be used for feature distillation. (d) After introducing new modalities, we not only adopted the feature distillation strategy of similar modalities, but also directly processed the interactive features between different modalities.}
    \label{fig:alignment}
\end{figure}

\section{Proposed method}
\label{sec:3}
\begin{figure*}
    \centering
    \includegraphics[width=\linewidth]{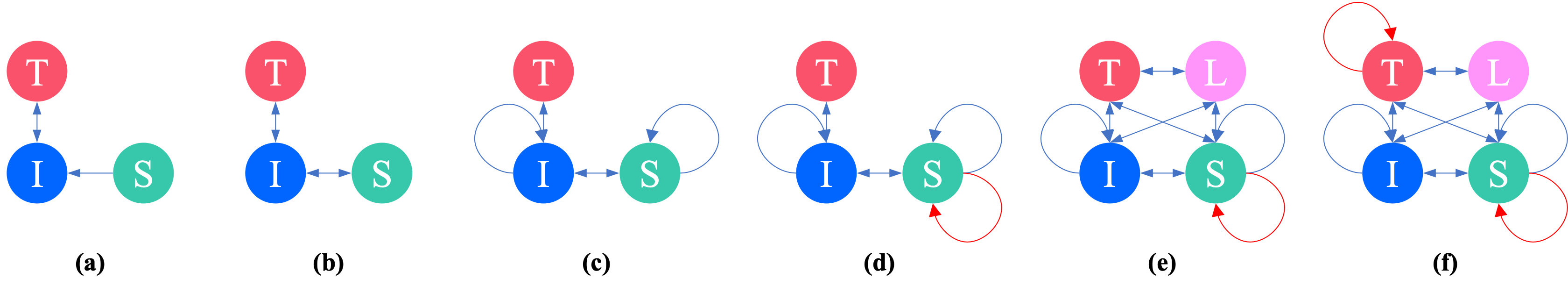}
    \caption{\textbf{Modality Graphs.} T=text feature, I=image feature, S=segmentation feature of NeRF, L=label volume feature of NeRF. (a) Previous methods only distilled the image modality capabilities of CLIP into the segmentation branch of NeRF. (b) and (c) demonstrate our attempt to align the segmentation modality of NeRF with the image modality of CLIP, as discussed in \Cref{sec:3.1}, where we introduced an adapter and a self-cross training strategy. (d) describes the self-enhancement of the NeRF segmentation modality, where in \Cref{sec:3.2}, we proposed a low-rank transient query attention. (e) By incorporating the label modality of NeRF, we achieved bilateral alignment among four modalities, as shown in \Cref{sec:3.3}, introducing label volume and $\mathcal{L}_{CE}$ loss. (f) pertains to the self-augmentation of the textual modality, where in \Cref{sec:3.4}, we proposed a simplified text augmentation strategy.}
    \label{fig:modality}
\end{figure*}

Previous work essentially treated NeRF features as new modalities to be fused with existing multi-modal models, aiming to handle downstream tasks such as segmentation and detection. The integration of this new modality with the multimodal model can be achieved in several different ways as shown in \Cref{fig:alignment}. 3D language-guided segmentation methods achieved cross-modal alignment by refining CLIP image features into the segmentation branch of NeRF, i.e. \Cref{fig:alignment} (c). However, these methods adopt a basic and rough alignment strategy and fail to deeply explore and utilize the potential of this alignment. Therefore, they will use post-processing methods for optimization, including pixel-level segmentation strategies and DINO. Our goal is to simplify these post-processing methods and utilize more refined alignment strategies to achieve efficient language-guided segmentation (\Cref{fig:alignment} (d)). This work aims to embed different modalities into the same vector space, so that in this space, semantic level comparison can be performed through a simple dot product operation. Given that three different modalities are involved: image, text, and NeRF segmentation features, these modal spaces can exist in multiple configurations of interrelationships. Alignment is achieved between text and image modalities, so we can exploit this relationship to implement language-based 3D segmentation guidance, as shown in \Cref{fig:modality}.

%Our pipeline is shown as \Cref{fig:architecture}.
Specifically, we propose Laser, a novel language-guided segmentation method reconstructing NeRF. By leveraging multi-view images of a given scene and open vocabulary descriptions for each category to segment the reconstructed NeRFs such that each 3D point is assigned a corresponding category label. To achieve this, we map each 3D point in the scene to CLIP features that represent its semantics. As CLIP only can obtain the image-level features, we use a modified CLIP encoder to derive dense image embeddings and learn an adapter for reducing the noise, as described in \Cref{sec:3.1}. However, variations in lighting and color in images can lead to segmentation ambiguity, resulting in inaccurate correlation values and misclassification. Taking a pair of shoes captured in a photo as an example, the corresponding CLIP features encompass the characteristics of the shoes. If the shoes have an all-black appearance, the shoe label is in a light black shade with a white logo, and the shoes are placed on a table with an all-white color and black stripes, CLIP features may mistakenly classify the shoe label as part of the table. We introduce a novel low-rank transient query attention and a simplified text augmentation strategy to enhance semantic features, as described in \Cref{sec:3.2} and \Cref{sec:3.4}. In addition, to ensure the consistency of segmentation maps among different views in the same scene, we introduced label volume. By transforming the segmentation problem into a classification task, we assign a corresponding category label to each 3D point, thereby enhancing the consistency of segmentation maps across different views, as described in \Cref{sec:3.3}.

\subsection{Mapping 3D to Dense Pixel-Level CLIP Features}
\label{sec:3.1}
To improve efficiency, we use the feature maps obtained by the recent CLIP dense prediction method \cite{wang2023sclip,ding2023maskclip} instead of the multi-scale CLIP feature maps used by previous methods \cite{lerf2023,liu2023weakly}. After obtaining the dense pixel-level CLIP features, we follow \cite{liu2023weakly,lerf2023,kobayashi2022decomposing}, introducing an additional branch to render the CLIP feature. We can render the RGB value and the CLIP feature of each ray $\textbf{r}$ using volume rendering \cite{mildenhall2021nerf}:
\begin{equation}
\hat{C} (\textbf{r}) = \sum_{i=1}^{N} w_iC_i\in \mathbb{R}^3 ,\hat{F} (\textbf{r}) = \sum_{i=1}^{N} w_iF_i\in \mathbb{R}^D,
\end{equation}
\begin{equation}
\label{wi}
\text{where} \quad w_i=\text{exp}\left ( -\sum_{j=1}^{i-1}\sigma_j\delta _j \right ) (1-\text{exp}(-\sigma_i\delta _i)),
\end{equation}
where $\sigma_i$ and $\delta_i$ refer to the volume density and feature attributes of the sampled point $i$, with $w_i$ indicating the weighting of $F_i$ within the ray $\textbf{r}$, and $\delta_i$ denoting the distance between adjacent samples. In every training batch, the supervision loss for a set of rays (denoted as $\mathcal{R}$) is defined as the sum of the L2 distance between the rendered and ground truth RGB values, and the cosine similarities $cos\left \langle , \right \rangle $ between the rendered features and the dense pixel-level CLIP features $F(\textbf{r})$:
\begin{equation}
\mathcal{L}_s=\sum_{\textbf{r}\in\mathcal{R}}\left ( \left \|\hat{C}(\textbf{r})-C(\textbf{r})\right \|_2 -cos\left \langle \hat{F}(\textbf{r}), F(\textbf{r})\right \rangle \right ).
\end{equation}

% \begin{figure}
%     \centering
%     \includegraphics[width=\linewidth]{adapter.png}
%     \caption{Caption}
%     \label{fig:adapter}
% \end{figure}

After training, given a set of text descriptions $\mathcal{T}  = \left \{ t_i \right \} _{i=1}^{N}$ and a CLIP text encoder $E_t$,  we can get the classes’ text features $F_t=E_t(\mathcal{T})\in \mathbb{R}^{N\times D}$. Then we can calculate the cosine similarity between the rendered CLIP features $\hat{F} (\textbf{r})$ and the class' text features $F_t$ to obtain the segmentation logits $z(\textbf{r})$ of the ray $\textbf{r}$:
\begin{equation}
z(\textbf{r}) = cos\left \langle \hat{F} (\textbf{r}), F_t\right \rangle \in \mathbb{R}^N.
\end{equation}
We can then get the segmentation maps of the ray $s(\textbf{r})=\text{argmax}(z(\textbf{r}))$. Similarly, we can get the segmentation logits $z_I(\textbf{r})$ of dense pixel-level CLIP features:
\begin{equation}
z_I(\textbf{r}) = cos\left \langle F(\textbf{r}), F_t\right \rangle \in \mathbb{R}^N.
\end{equation}

Although using dense pixel-level CLIP feature improves efficiency, it also includes a lot of noise. To address this issue, we present a novel adapter $f_{\theta a}$ that includes two components, an attention module and a small bottleneck layer. We aim to reconstruct the dense CLIP feature enabling it to reduce noise:
\begin{equation}
F(\textbf{r})_r \approx f_{\theta a}(F(\textbf{r})),
\end{equation}
we can then use the cosine similarity to optimize $f_{\theta a}$:
\begin{equation}
\mathcal{L}_r = cos\left \langle F(\textbf{r})_r, (F(\textbf{r}) \right \rangle.
\label{adapter_opm_org}
\end{equation}
The new dense pixel-level CLIP feature captured is added with the original features via residual connections:
\begin{equation}
F(\textbf{r})_r^{\text{new}} = \alpha F(\textbf{r})_r + (1-\alpha)F(\textbf{r}).
\end{equation}

However, this optimization is ineffective because we have not made any fine-tuning of CLIP. We observe that after a certain number of training steps, the rendered CLIP features are difficult to improve due to the limitations of CLIP features. Additionally, the rendered CLIP features are cleaner than CLIP features but sometimes lack some semantic information. Therefore, we propose a self-cross-training strategy that aims to reduce noise and improve rendered CLIP features by optimizing adapters to achieve better performance. To sum up, the $f_{\theta a}$ can be found via solving:
\begin{equation}
\begin{split}
&\mathcal{L}_r =  - \beta\left(\frac{f_{\theta a}(F(\textbf{r})) \cdot F(\textbf{r})}{\left | f_{\theta a}(F(\textbf{r})) \right | \left | F(\textbf{r}) \right | } + \frac{f_{\theta a}(\hat{F}(\textbf{r})) \cdot \hat{F}(\textbf{r})}{\left | f_{\theta a}(\hat{F}(\textbf{r})) \right | \left | \hat{F}(\textbf{r}) \right | } \right)\\
&-(1-\beta)\left( \frac{f_{\theta a}(\hat{F}(\textbf{r})) \cdot F(\textbf{r})}{\left | f_{\theta a}(\hat{F}(\textbf{r})) \right | \left | F(\textbf{r}) \right | } 
+ \frac{f_{\theta a}(F(\textbf{r})) \cdot \hat{F}(\textbf{r})}{\left | f_{\theta a}(F(\textbf{r})) \right | \left | \hat{F}(\textbf{r}) \right | }\right),
\end{split}
\label{adapter_opm_sct}
\end{equation}
where $\beta$ is a weight, and we find through experiments that setting it to $0.3$ is optimal. The noise can be mitigated by optimizing adapters and the rendered CLIP feature is cleaner than the original CLIP feature, but its semantic information at the edges is blurred.

\subsection{Low-Rank Transient Query Attention}
\label{sec:3.2}
For the problem of blurred edges of rendered CLIP features, an intuitive idea is to utilize convolution or attention mechanisms, applying them directly to rendered CLIP features may result in a loss of spatial information \cite{lerf2023}. Hence, we try to enhance these features by finding relationships between 3D points. There are some challenges in using convolution or attention mechanisms directly on 3D points. First of all, due to possible occlusion problems in space, the 3D points participating in the training are not fixed at each step. Secondly, the large number of 3D points leads to a significant increase in complexity, which is unacceptable. To address this challenge, we consider the use of low-rank decomposition to streamline the computation within attention mechanisms\cite{zhou2014low}. By reducing the rank of the matrix, this approach effectively reduces computing complexity and storage requirements, enabling representation through a reduced number of parameters. In attention mechanisms, this allows for an approximation of the original attention matrix by a lower-dimensional equivalent, preserving essential information with minimal loss. The primary benefit of low-rank decomposition lies in its efficiency in cutting down computational demands, a crucial factor when dealing with extensive data sets. By breaking down the initial high-dimensional attention matrix into two smaller matrices, we effectively lessen both the size and computational load of the model, whilst retaining its critical features.

Given a batch of sampling points, we can obtain their corresponding features from the feature grid, represented as $F_{sp}\in\mathbb{R}^{S\times D}$, $S$ is the number of sampling points in a batch (i.e., the product of the number of rays and the number of sampling points along each ray), and $D$ is the number of feature channels. For the vanilla self-attention mechanism:
\begin{equation}
Q=q(F_{sp}),K=k(F_{sp}),V=v(F_{sp}),
\end{equation}
where $q,k,v$ represent $1\times 1$ convolution layers. We can then get the $\hat{F}_{sp}$:
\begin{equation}
\hat{F}_{sp}=V\otimes \text{Softmax}(Q\otimes K).
\end{equation}
The complexity can be expressed as $\mathcal{O}(S^2D)$, where $S$ is a large number, typically exceeding 20,000. Consequently, employing vanilla attention with a global receptive field results in overwhelming computation complexity. 

\begin{figure}
    \centering
    \includegraphics[width=0.9\linewidth]{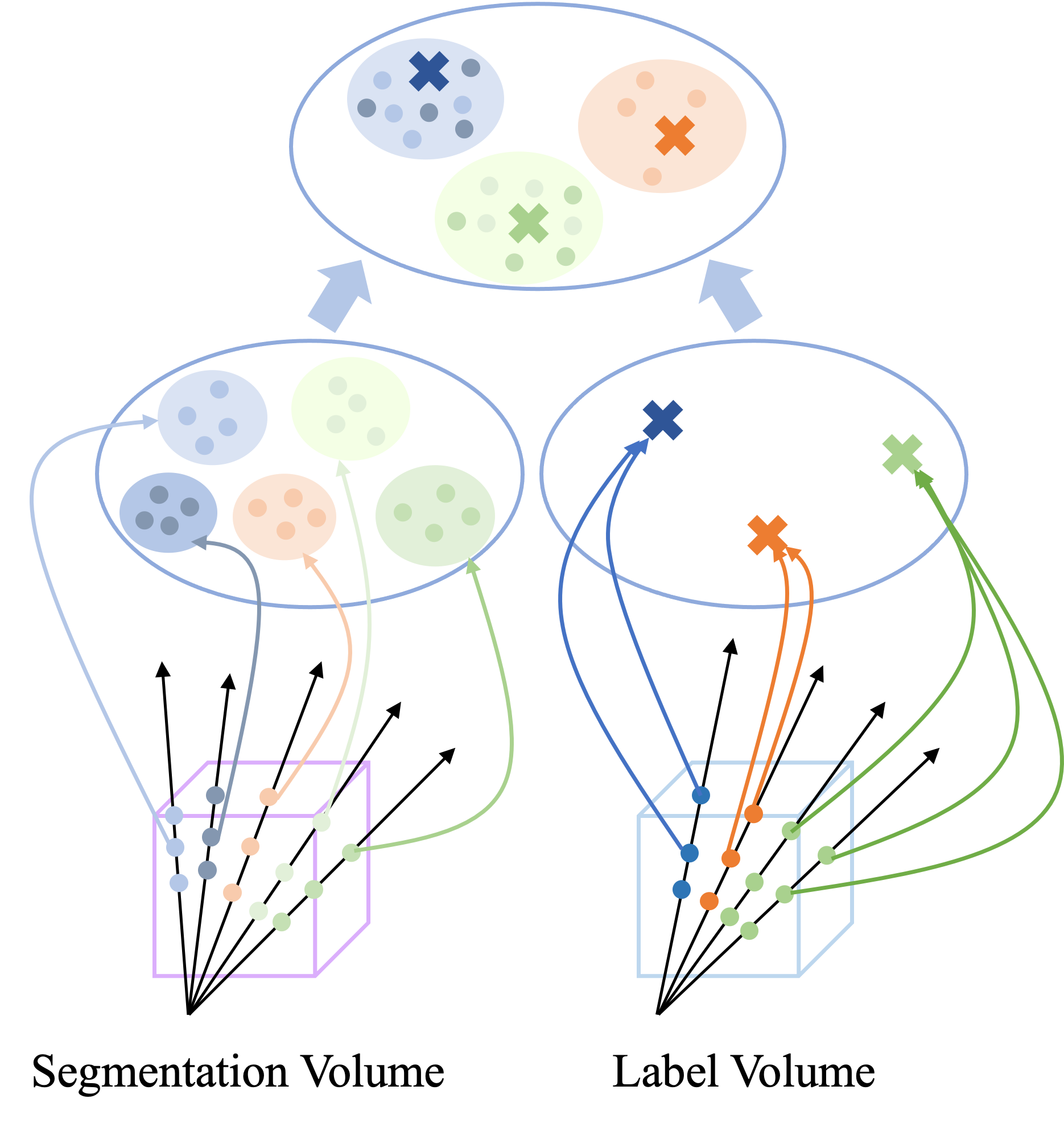}
    \caption{\textbf{Employing label volume to generate cluster centroids.} Our method progressively aggregates points lying on the same ray into a shared cluster centroid during training. This process effectively groups 3D points, which are spatially represented on similar-looking features, into the same category. As a result, 3D points that share close appearances in their feature are associated with the same cluster, reinforcing the consistency of the categorization based on their color similarities.}
    \label{fig:center}
\end{figure}

For a matrix $W\in\mathbb{R}^{d\times k}$, we represent it by a low-rank decomposition $W=BA$, where $B\in\mathbb{R}^{d\times r}, A\in\mathbb{R}^{r\times k}$, and the rank $r \ll min(d,k)$. This decomposition allows us to approximate the attention matrix efficiently with reduced dimensions. Thus, we propose a novel low-rank transient query attention, leveraging a transient query learnable parameter $T\in\mathbb{R}^{s\times D}$.
For vanilla attention output $\hat{F}_{sp}$, we can regard it as $W_F\in\mathbb{R}^{S\times D}$ and factorized it:
\begin{equation}
\begin{aligned}
W_F &= \underbrace{\text{M}_{qk} \circ \text{v}_v}_{\text{Vanilla Attention}}  \\
    &= \underbrace{\text{M}_{qT}\circ \text{M}_{kT} \circ \text{v}_v}_{\text{Transient Query Attention}}  \\
\end{aligned}
\end{equation}
where $\text{M}_{qk}\in\mathbb{R}^{S\times S}$, $\text{M}_{qT}\in\mathbb{R}^{S\times s}$, $\text{M}_{kT}\in\mathbb{R}^{s\times S}$, and $\text{v}_v\in\mathbb{R}^{S\times D}$. The $\text{M}_{qT}$ and $\text{M}_{kT}$ can approximate the $\text{M}_{qk}$, and $s \ll min(S,D)$.
Specifically, we can use $T$ to query $Q$ and $K$ respectively and calculate Softmax:
\begin{equation}
\hat{T}_q=\text{Softmax}(T\otimes Q)\in\mathbb{R}^{s\times S},
\end{equation}
\begin{equation}
\hat{T}_k=\text{Softmax}(T\otimes K)\in\mathbb{R}^{s\times S}.
\end{equation}
Next, we apply $\hat{T}_k$ to query $V$ and calculate Softmax as the $V$ of vanilla attention:
\begin{equation}
\hat{V}=\text{Softmax}(\hat{T}_k\otimes V)\in\mathbb{R}^{s\times D}.
\end{equation}
We then can get the $\hat{F}_{sp}$:
\begin{equation}
\hat{F}_{sp}=\hat{T}_q\otimes \hat{V}\in\mathbb{R}^{S\times D}.
\end{equation}
Consequently, the complexity is $\mathcal{O}(SsD)$. This reduction in complexity significantly contributes to a more efficient training process, thereby effectively minimizing training time.

\begin{table*}
\caption{\textbf{Quantitative comparison} across competing methods on the 3D-OVS \cite{liu2023weakly}, where overall indicates the average results across all the metrics in the three different scenes. We highlight the \colorbox{best}{best}, \colorbox{second}{second-best}, and \colorbox{third}{third-best} scores.}
\label{tab:sota}
\resizebox{\textwidth}{!}{
\begin{tabular}{c|c|cc|cc|cc|cc|cc|cc|cc}
\toprule
\multirow{2}{*}{Methods} & \multirow{2}{*}{\shortstack{Training\\Time}} & \multicolumn{2}{c|}{\textit{bed}} & \multicolumn{2}{c|}{\textit{sofa}} & \multicolumn{2}{c|}{\textit{lawn}} & \multicolumn{2}{c|}{\textit{room}} & \multicolumn{2}{c|}{\textit{bench}} & \multicolumn{2}{c|}{\textit{table}}& \multicolumn{2}{c}{\textit{overall}} \\ \cline{3-16} 
                        &                                & mIoU     & Acc.     & mIoU      & Acc.     & mIoU      & Acc.     & mIoU      & Acc.     & mIoU      & Acc.      & mIoU      & Acc.      & mIoU      & Acc.  \\ \hline
LSeg \cite{li2022languagedriven}                    & -                              & 56.0     & 87.6         & 04.5      & 16.5         & 17.5      & 77.5         & 19.2      & 46.1         & 06.0      & 42.7          & 07.6      & 29.9   & 18.5      & 50.1\\
ODISE \cite{xu2023open}                   & -                              & 52.6     & 86.5         & 48.3      & 35.4         & 39.8      & 82.5         & 52.5      & 59.7         & 24.1      & 39.0          & 39.7      & 34.5  & 42.8      & 56.3\\
OV-Seg  \cite{liang2023open}                & -                              & \cellcolor{third}79.8     & 40.4         & \cellcolor{third}66.1      & \cellcolor{third}69.6         & \cellcolor{third}81.2      & 92.1         & \cellcolor{third}71.4      & 49.1         & \cellcolor{second}88.9      & \cellcolor{third}89.2          & \cellcolor{third}80.6      & \cellcolor{third}65.3  & \cellcolor{third}78.0      & 67.6\\

DFF \cite{kobayashi2022decomposing}                 & 184 min                        & 56.6     & \cellcolor{third}86.9         & 03.7      & 09.5         & 42.9      & 82.6         & 25.1      & 51.4         & 06.1      & 42.8          & 07.9      & 30.1  & 23.7      & 50.6\\
LERF \cite{lerf2023}                   & \cellcolor{second}54 min                         & 73.5     & \cellcolor{third}86.9         & 27.0      & 43.8         & 73.7      & \cellcolor{third}93.5         & 46.6      & \cellcolor{third}79.8         & 53.2      & 79.7          & 33.4      & 41.0  & 51.2      & \cellcolor{third}70.8\\
3D-OVS  \cite{liu2023weakly}                & \cellcolor{third}158 min                        & \cellcolor{second}89.5     & \cellcolor{second}96.7         & \cellcolor{second}74.0      & \cellcolor{second}91.6         & \cellcolor{second}88.2      & \cellcolor{second}97.3         & \cellcolor{best}92.8      & \cellcolor{best}98.9         & \cellcolor{best}89.3      & \cellcolor{second}96.3          & \cellcolor{best}88.8      & \cellcolor{best}96.5   & \cellcolor{second}87.1      & \cellcolor{second}96.2\\ \hline
Our                     & \cellcolor{best} 11 min                         & \cellcolor{best}91.4     & \cellcolor{best}98.0         & \cellcolor{best}86.0      & \cellcolor{best}96.2         & \cellcolor{best}88.5      & \cellcolor{best}98.4         &   \cellcolor{second}85.9        &     \cellcolor{second}97.7         &   \cellcolor{third}88.3        &     \cellcolor{best}96.9          & \cellcolor{second}88.5      & \cellcolor{second}96.3   & \cellcolor{best}88.1     & \cellcolor{best}97.3\\ \bottomrule
\end{tabular}
}
\end{table*}

\begin{figure}
    \centering
    \includegraphics[width=\linewidth]{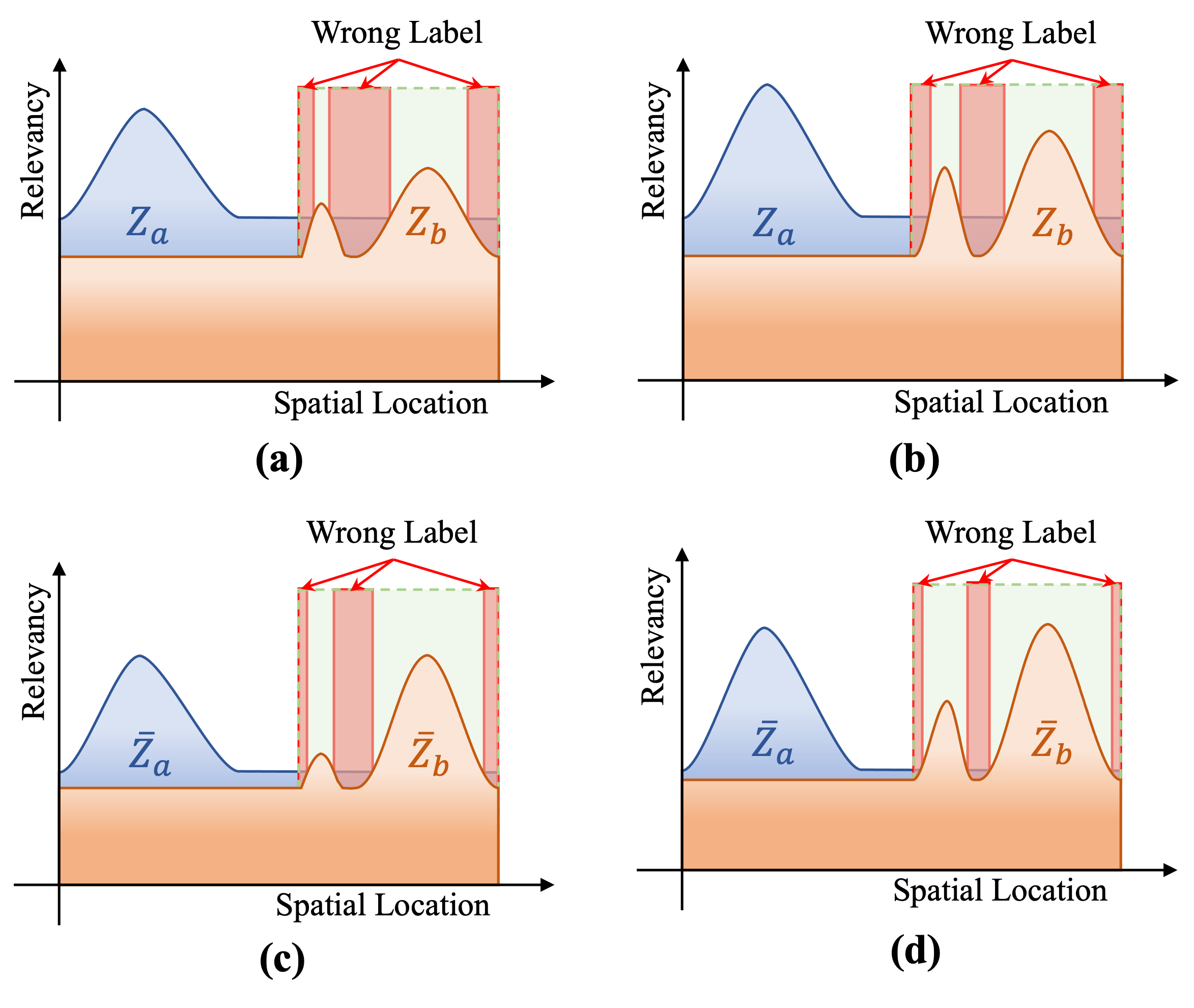}
    \caption{\textbf{Mitigating the ambiguity in CLIP features.} We employ a simplified text augmentation strategy to standardize relevance maps. Observing the original relevance maps $Z_a$ and $Z_b$ in (a), we note that the relevance of class \textbf{a} within the red-highlighted area is higher than in other image regions. Due to the higher absolute relevance of class \textbf{a} in this area, the ambiguity of CLIP features results in the red region being classified as class \textbf{a}, even though class \textbf{b} is also present. In (b), we reduce this ambiguity by simply repeating the text to recalculate the relevance maps $Z_a$ and $Z_b$, thereby enhancing the accuracy of regional class assignments. In (c), standardizing the relevance maps of each class to a fixed range also can reduce ambiguity. In (d), we combine text repetition with standardization of the relevance maps, significantly reducing the ambiguity in classification and leading to more precise regional class allocations. }
    \label{fig:relevancy}
\end{figure}
\begin{figure*}
    \centering
    \includegraphics[width=\linewidth]{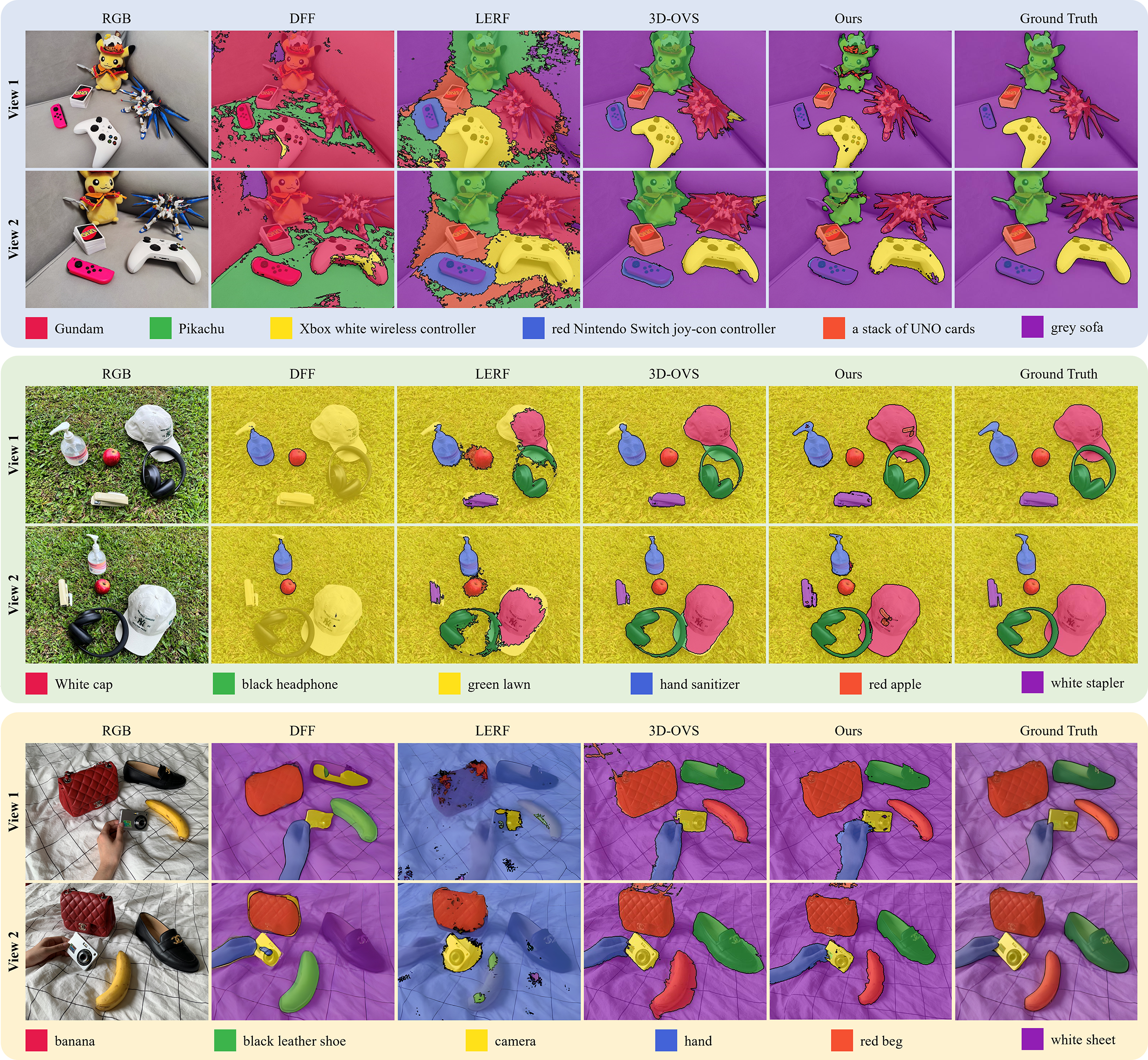}
    \caption{\textbf{Qualitative comparison.} Visualization of segmentation results in 3D-OVS dataset 3 scenes \cite{liu2023weakly}. Note that we generate object masks in this dataset requiring the complete category list.}
    \label{fig:qualitative}
\end{figure*}

\subsection{Label Volume for Enhance Consistency}
\label{sec:3.3}
In 3D scenes, maintaining consistent views is crucial. To this end, we propose a novel label volume to transform the segmentation problem into a classification problem. Specifically, each 3D point $\textbf{x}\in\mathbb{R}^3$ yields a label vector $L_\textbf{x}\in\mathbb{R}^3$ from label volume $L$. Then we can render the label vector of each ray $\textbf{r}$ using volume rendering:
\begin{equation}
L(\textbf{r})=\text{Softmax}(\sum_{i=1}^{N} w_iL_i)\in [0,1]^D.
\end{equation}
where $w_i$ refers to \Cref{wi}, we employ a Softmax function on the label vector for each ray, ensuring that the probabilities sum up to 1. We can calculate pseudo labels $L_{pGT}$ using classes’ text features $F_t$:
\begin{equation}
L_{pGT}=cos\left \langle L(\textbf{r}), F_t\right \rangle \in \mathbb{R}^N.
\end{equation}
We can then use cross-entropy to optimize $z_I(\textbf{r})$ and $z(\textbf{r})$:
\begin{equation}
\label{ce}
\mathcal{L}_{CE} = - \sum_{i=1}^{N} L_{pGT_{i}} \cdot \left( \log(z_I(\mathbf{r})_{i}) + \log(z(\mathbf{r})_{i}) \right),
\end{equation}
where $N$ represents the number of classes. As illustrated in \Cref{sec:3.1}, $z_I(\mathbf{r})$ include certain noise and $z(\textbf{r})$ may lose some semantic information. However, $L_{pGT}$ is a learnable smoothing pseudo label, during the backpropagation process, these noisy and semantically inaccurate areas can contribute to incorrect gradients, leading to inaccuracies in labels, as illustrated in \Cref{fig:center}. To alleviate this issue, we adopt the concept of ensemble learning, weighting $z_I(\mathbf{r})$ and $z(\textbf{r})$, and optimizing the smoothed pseudo-label $L_{pGT}$. The expression for \Cref{ce} can be expressed as follows:
\begin{equation}
\begin{split}
\mathcal{L}_{CE} = - \sum_{i=1}^{N} & L_{pGT_{i}} \cdot \left( \log(z_I(\mathbf{r})_{i}) + \log(z(\mathbf{r})_{i}) \right.\\
& \left. + \log(\gamma z_I(\mathbf{r})_{i} + (1-\gamma) z(\mathbf{r})_{i}) \right),
\end{split}
\end{equation}
where $\gamma$ control $z_I(\mathbf{r})$ and $z(\textbf{r})$ weight. By minimizing $L_{sGT}$, we ensure that 3D points on each ray in the label volume are aggregated onto their most probable category, causing the 3D point set on the segmentation branch to concentrate on relatively correct classifications. Relatively accurate classification of a given 3D point ensures consistency across views during rendering. Although converting the classification problem can obtain relative accuracy segmentation, it cannot handle the ambiguity between original CLIP features and textual correlations, as illustrated in \Cref{fig:relevancy}.

\subsection{Simplified Text Augmentation for Mitigating Ambiguity}
\label{sec:3.4}
To alleviate the ambiguity between original CLIP features and textual correlations, we propose a simplified text augmentation strategy. Specifically, we let the text descriptions $\mathcal{T}$ repeat twice to obtain $\mathcal{T}_{aug}  = \left \{ t_i \times 2\right \} _{i=1}^{N}$, and then get the classes’ text features $F_t^{aug}=E_t(\mathcal{T}_{aug})\in \mathbb{R}^{N\times D}$. Given an image, we can get it dense pixel-level CLIP feature $F_I\in \mathbb{R}^{H\times W\times D}$ from the CLIP image encoder and though the adapter. We can then get the image segmentation logits $z_I$:
\begin{equation}
z_I = cos\left \langle F_I, F_t^{aug}\right \rangle \in \mathbb{R}^{N\times H\times W}.
\end{equation}
We independently normalize the correlation of each class in the input view to $[0, 1]$ to alleviate the ambiguity of CLIP features:
\begin{equation}
\bar{z}_I =\frac{z_I-\text{min}(z_I)}{\text{max}(z_I)-\text{min}(z_I)} \in[0,1]^{N\times H\times W}.
\end{equation}
Here, the functions $\text{min}(\cdot)$ and $\text{max}(\cdot)$ are utilized to retrieve the minimum and maximum values within the spatial dimensions (i.e. $H$ and $W$). We can then further optimize the segmentation logits $z_I(\textbf{r})$ of the ray $\textbf{r}$:
\begin{equation}
\mathcal{L}_{aug}=cos\left \langle z_I(\textbf{r}), \bar{z}_I(\textbf{r}) \right \rangle.
\end{equation}
Note that $\bar{z}_I(\textbf{r})$ is to select the image pixels corresponding to the batch of rays $\textbf{r}$.

\section{Experiment}
\subsection{Implementation Details}
To reconstruct a 3D scene from multiview images, we follow the same training settings as TensoRF \cite{chen2022tensorf} and 3D-OVS \cite{liu2023weakly}. During the segmentation phase of training, we train the model for 15,000 iterations. For the first 5,000 iterations, we freeze the parameters of the appearance volume and density volume and train only on the label volume, CLIP feature branches, and adapters. In the subsequent 10,000 iterations, we further fine-tune the appearance volume and RGB branches. At the 8,000th iteration, we introduce a self-crossing loss function to optimize the adapter. We used the Adam optimizer with \textit{betas} = (0.9, 0.99). The initial learning rates used to train the volumetric and segmentation MLP branches are 0.02 and 1\textit{e}-4 respectively. During the fine-tuning phase, these learning rates are adjusted to 5\textit{e}-3 and 5\textit{e}-5. In addition, we also implemented a learning rate decay strategy with a decay coefficient of 0.1. The model dynamically computes dense pixel-level CLIP features for each training view during training. We use the ViT-B/16 CLIP model to extract the image and text features. The model runs on an NVIDIA 3090 GPU with 24G of memory. The average training time is approximately 11 minutes, and the memory required during training does not exceed 10G.

\begin{figure}
    \centering
    \includegraphics[width=\linewidth]{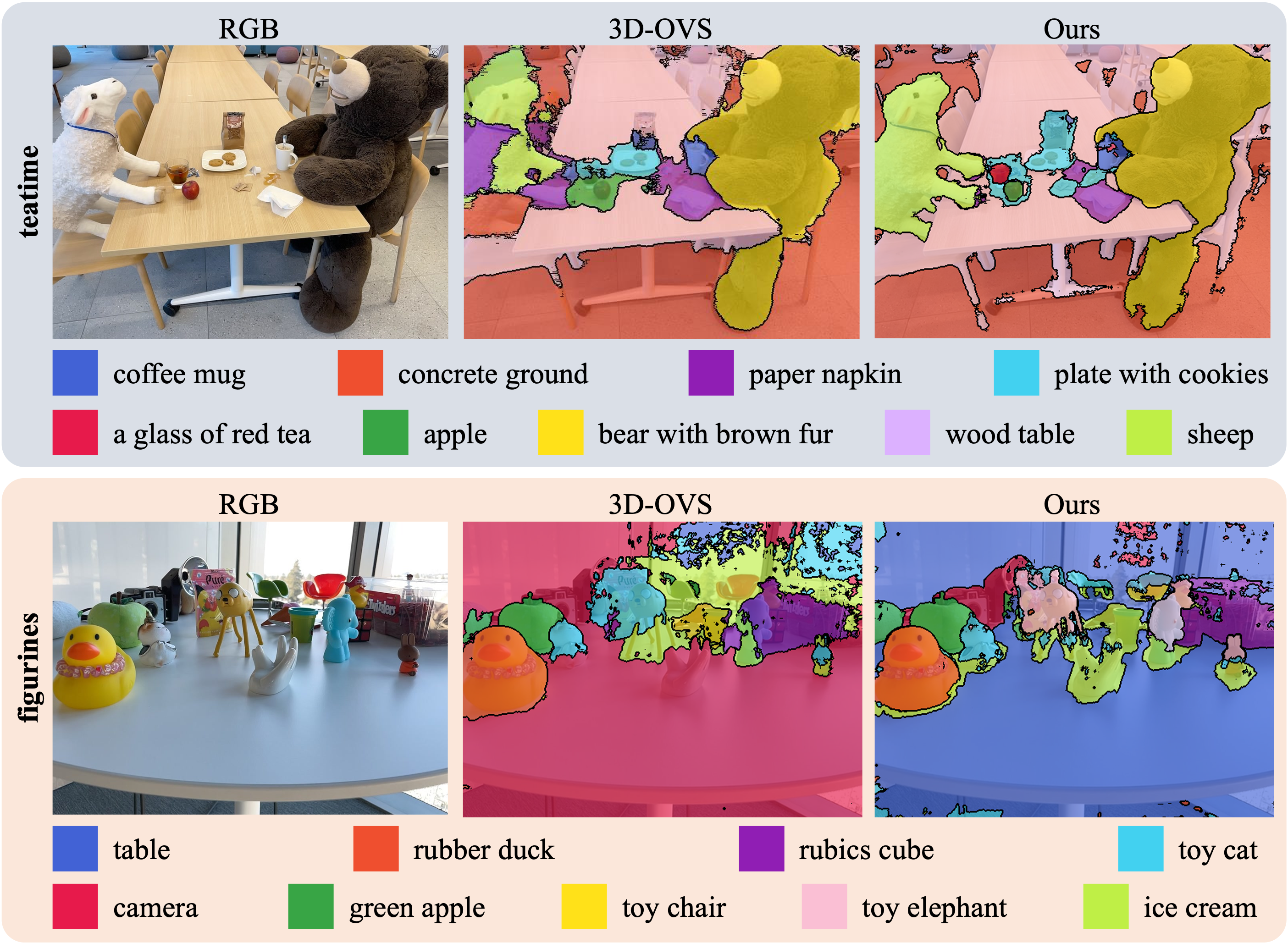}
    \caption{\textbf{Qualitative comparison.} Visualization of segmentation results in LERF \cite{lerf2023} dataset.}
    \label{fig:qualitative_1}
\end{figure}

\begin{figure}
    \centering
    \includegraphics[width=\linewidth]{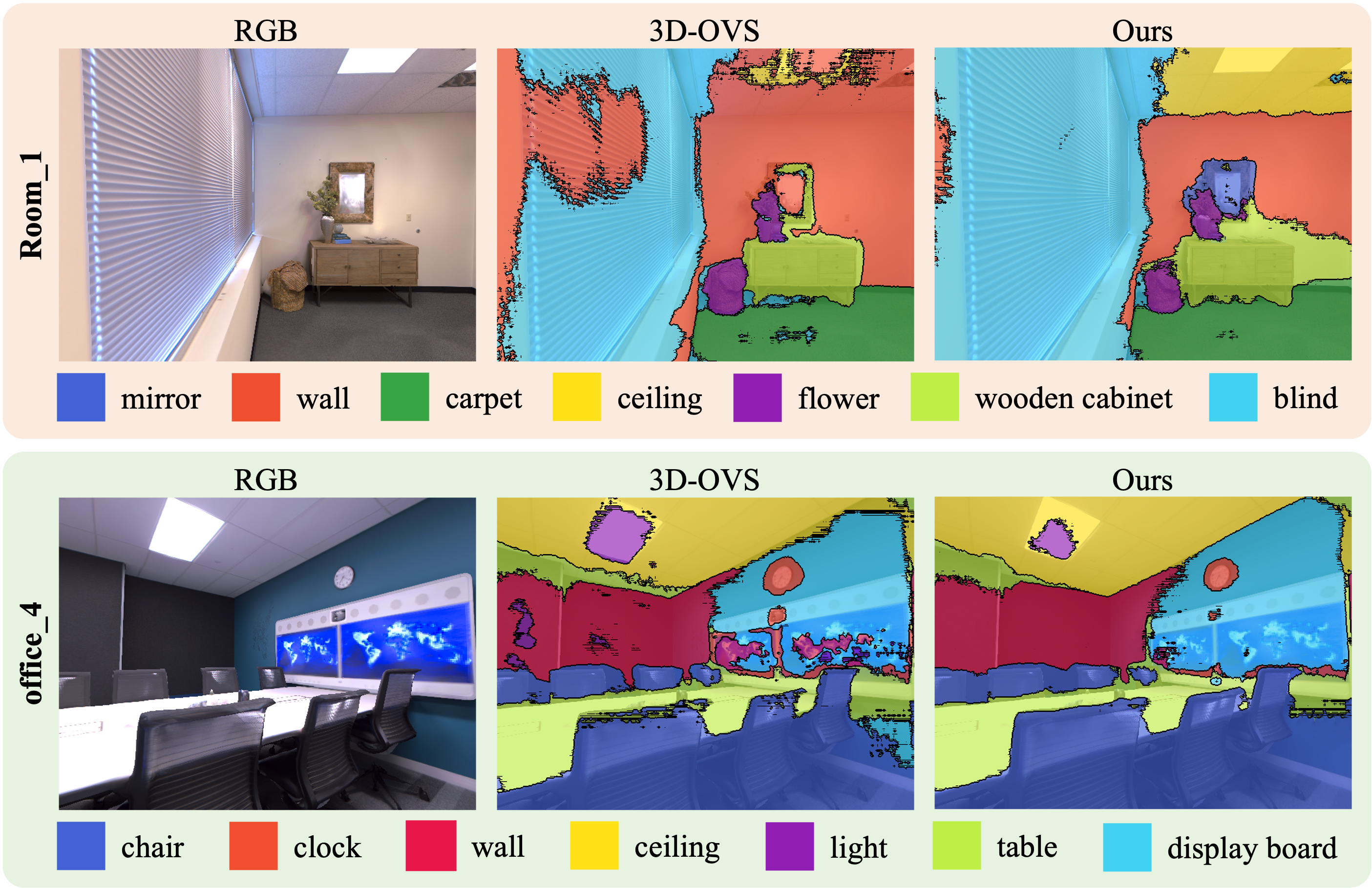}
    \caption{\textbf{Qualitative comparison.} Visualization of segmentation results in Replica \cite{straub2019replica} dataset.}
    \label{fig:qualitative_2}
\end{figure}

\subsection{Evaluation Metrics}
In this work, we adopt two metrics, the mIoU score and the accuracy score, to evaluate the performance of our image segmentation model. mIoU score is a standard indicator to measure the segmentation effect of the model. It calculates the average of the ratio of intersection and union between the segmentation area predicted by the model and the true segmentation area. The accuracy score measures the proportion of pixels correctly classified by the model, that is, the number of pixels correctly predicted by the model as a proportion of the total number of pixels.

\subsection{Dataset}
To assess the efficacy of our approach, we primarily focus on the 3D-OVS \cite{liu2023weakly} dataset, which is specially designed for open-vocabulary 3D semantic segmentation and features a diverse collection of long-tail objects in various poses and backgrounds, complete with a detailed list of categories. In addition to the 3D-OVS dataset, we further validate our method using two supplementary datasets. The first is the widely-used Replica \cite{straub2019replica} dataset, targeting multi-view indoor scenes. The second is the LERF \cite{lerf2023} dataset, which is derived from complex real-world scenes captured via the iPhone App Polycam. 

\subsection{Inference Cost}
We compared the inference time and FPS between our method and 3D-OVS \cite{liu2023weakly}. As shown in \Cref{tab:inf_time}, due to the retention of the adapter and low-rank transient query attention module in the inference phase, our inference time and FPS have decreased. However, it should be emphasized that the main focus of this work is to reduce the consumption of the training phase rather than the consumption of the inference phase.
\begin{table}
\centering
\caption{\textbf{Comparison of inference time and FPS} for our and 3D-OVS.}
\begin{tabular}{ccc}
\toprule
Method & Inference Time (s/Frame) & FPS    \\ \midrule
3D-OVS & 6.156 & 0.16 \\ 
Our    & 9.125 & 0.11 \\ \bottomrule
\end{tabular}
\label{tab:inf_time}
\end{table}

\subsection{Comparison with State-of-the-Art Methods}
\subsubsection{Qualitative Results}
In Figures \Cref{fig:qualitative}, \Cref{fig:qualitative_1}, and \Cref{fig:qualitative_2}, we present the visualization results of three datasets. For the 3D-OVS \cite{liu2023weakly} dataset, we demonstrate the segmentation effects observed from different perspectives, as shown in \Cref{fig:qualitative}. Compared to other 3D methods, our methods demonstrate superior performance in capturing fine edge details. For the LERF \cite{lerf2023} and Replica \cite{straub2019replica} datasets, we compared our method with the current state-of-the-art 3D open vocabulary segmentation method, 3D-OVS. In terms of processing complex scenes, our method can more precisely identify scene details, such as the apples and teacups in the teatime scene, and the mirror in the room\_1 scene. These qualitative results prove the advantages of our proposed method in generating accurate object boundaries and ensuring 3D semantic consistency.

\subsubsection{Quantitative Results}
\begin{table*}
\caption{\textbf{Quantitative comparison} across competing methods on the LERF \cite{lerf2023} and Replica \cite{straub2019replica}. We highlight the \colorbox{best}{best} scores.}
\label{tab:sota_1}
\resizebox{\linewidth}{!}{
\begin{tabular}{c|cccccccc|cccccccc}
\toprule
\multirow{3}{*}{Methods}                     & \multicolumn{8}{c|}{LERF \cite{lerf2023}}                                                                                                                                                                                                                                                                                                                                                                        & \multicolumn{8}{c}{Replica \cite{straub2019replica}}                                                                                                                                                                                                                                                                                                                                                                         \\ \cline{2-17} 
                                             & \multicolumn{2}{c|}{\textit{teatime}}                                                                       & \multicolumn{2}{c|}{\textit{figurines}}                                                                     & \multicolumn{2}{c|}{\textit{ramen}}                                                                         & \multicolumn{2}{c|}{\textit{waldo\_kitchen}}                                           & \multicolumn{2}{c|}{\textit{room\_0}}                                                                       & \multicolumn{2}{c|}{\textit{room\_1}}                                                                       & \multicolumn{2}{c|}{\textit{office\_3}}                                                                     & \multicolumn{2}{c}{\textit{office\_4}}                                                 \\ \cline{2-17} 
                                             & mIoU                                  & \multicolumn{1}{c|}{Acc.}                                  & mIoU                                  & \multicolumn{1}{c|}{Acc.}                                  & mIoU                                  & \multicolumn{1}{c|}{Acc.}                                  & mIoU                                  & Acc.                                  & mIoU                                  & \multicolumn{1}{c|}{Acc.}                                  & mIoU                                  & \multicolumn{1}{c|}{Acc.}                                  & mIoU                                  & \multicolumn{1}{c|}{Acc.}                                  & mIoU                                  & Acc.                                  \\ \hline
LERF \cite{lerf2023}        & 26.0                                  & \multicolumn{1}{c|}{82.8}                                  & 34.6                                  & \multicolumn{1}{c|}{55.0}                                  & 25.2                                  & \multicolumn{1}{c|}{68.3}                                  & 33.9                                  & 42.7                                  & 7.7                                   & \multicolumn{1}{c|}{38.2}                                  & 12.8                                  & \multicolumn{1}{c|}{29.7}                                  & 12.6                                  & \multicolumn{1}{c|}{37.2}                                  & 8.2                                   & 16.3                                  \\
3D-OVS \cite{liu2023weakly} & 56.1                                  & \multicolumn{1}{c|}{87.7}                                  & 44.8                                  & \multicolumn{1}{c|}{77.3}                                  & 28.7                                  & \multicolumn{1}{c|}{70.2}                                  & 39.3                                  & 45.6                                  & 12.5                                  & \multicolumn{1}{c|}{40.1}                                  & 13.1                                  & \multicolumn{1}{c|}{30.2}                                  & 12.1                                  & \multicolumn{1}{c|}{36.1}                                  & 15.8                                  & 28.9                                  \\ 
Our & \cellcolor{best}62.4 & \cellcolor{best}88.0 & \cellcolor{best}63.5 & \cellcolor{best}83.9 & \cellcolor{best}44.6 & \cellcolor{best}75.8 & \cellcolor{best}41.3 & \cellcolor{best}48.9 & \cellcolor{best}24.3 & \cellcolor{best}54.7 & \cellcolor{best}25.0 & \cellcolor{best}45.9 & \cellcolor{best}18.7 & \cellcolor{best}64.3 & \cellcolor{best}39.7 & \cellcolor{best}62.5 \\ \bottomrule
\end{tabular}
}
\end{table*}

In \Cref{tab:sota}, we benchmark our method using three 2D open vocabulary segmentation methods \cite{li2022languagedriven,liang2023open,xu2023open} and three NeRF-based methods capable of 3D open vocabulary segmentation: DFF \cite{kobayashi2022decomposing}, LERF \cite{lerf2023}, and 3D-OVS \cite{liu2023weakly}. DFF is a pioneering work in the field of 3D open vocabulary segmentation, which applies the feature maps of the 2D open vocabulary segmentation method LSeg \cite{li2022languagedriven} to achieve its segmentation results. LERF, 3D-OVS are closely aligned with our proposed methods as they use knowledge distillation from CLIP and DINO. We observe that our method significantly surpasses existing 2D and 3D methods in multiple aspects. Our method not only outperforms 2D-based methods such as ODISE \cite{xu2023open} and OV-Seg \cite{liang2023open} in performance, but also outperforms 3D-based methods including LERF \cite{lerf2023} and 3D-OVS \cite{liu2023weakly}. Specifically, our method performs better than the current state-of-the-art 3D-OVS in 3 out of 6 scenes in the 3D-OVS dataset, and is comparable to 3D-OVS in 1 scene. In particular, significant performance improvements are achieved in the sofa scene. Although 3D-OVS adopts the optimization strategy of multi-scale pixel-level CLIP features and DINO features, the overall performance of our method is outstanding in terms of average mIoU score and accuracy. In addition, from an efficiency perspective, 3D-OVS consumes a lot of resources during the training phase, and the optimization of a single scene takes 158 minutes. In comparison, our method only takes 11 minutes for single-scene optimization, and is 14.3 times more efficient than 3D-OVS. In \Cref{tab:sota_1}, we compare our method against LERF \cite{lerf2023} and 3D-OVS \cite{liu2023weakly} using the LERF \cite{lerf2023} and Replica \cite{straub2019replica} datasets. Our method consistently demonstrates superior performance across the 8 scenarios evaluated. This quantitative experiment demonstrates our method's robust capability in executing precise NeRF-based 3D word segmentation, relying on a streamlined architecture.
\begin{figure*}
    \centering
    \includegraphics[width=\linewidth]{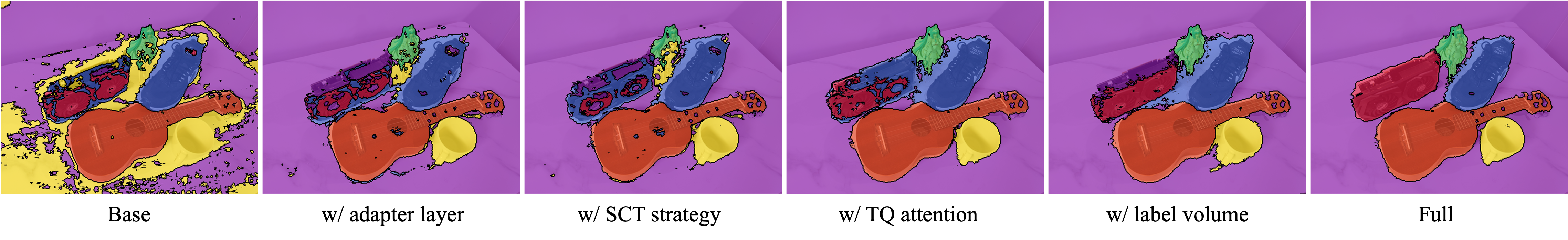}
    \caption{\textbf{Ablation studies visualization results} on the table scene of 3D-OVS \cite{liu2023weakly}. We present the visualization results of the adapter layer, the label volume, the self-cross-training (SCT) strategy, the low-rank transient query (TQ) attention, and the full module.}
    \label{fig:ablation_main}
\end{figure*}

\subsection{Analysis}
In this subsection, we performed an ablation study for different components of our Laser framework. In \Cref{tab:ab_main}, we initially start distilling dense CLIP features to the segmentation branch as the base. Subsequently, we gradually introduce other components, demonstrating the changes in mIoU scores and accuracy scores with the addition of each component based on this foundation. Additionally, we also show the qualitative results of this change (the table scene of 3D-OVS), as shown in the \Cref{fig:ablation_main}.
\subsubsection{Investigation of Optimizing Adapter Layer} We studied the importance of the designed components in our Laser framework. Specifically, we define the following methods: \emph{A)} \textit{w/ adapter layer} to directly optimize it use  \Cref{adapter_opm_org}, and \emph{B)} \textit{w/ SCT strategy} to directly evaluate our self-cross training strategy (\Cref{adapter_opm_sct}). The first three columns of  \Cref{fig:ablation_main} show the comparison of the adapter layer, self-cross training strategy with the baseline. From the results, although the segmentation effect can be achieved by directly extracting pixel-level features from distilled dense CLIP, since CLIP itself only has classification functions and lacks segmentation capabilities, a large amount of noise will be generated if the segmentation task is not specifically processed, refer to \Cref{sec:3}. \emph{Method A} can effectively reduce noise, thanks to the features reconstructed through the bottleneck layer, resulting in cleaner output features. \emph{Method B} can further reduce noise. The effectiveness of this strategy is based on our observation that rendered features demonstrate higher clarity compared to reconstructed features. Therefore, by enabling mutual supervision between these two types of features, we facilitate their mutual enhancement. Furthermore, we delved into the significance of the adapter residual connection ratio $\alpha$ and the ratio $\beta$ in the self-cross training strategy on the importance to the model. In \Cref{tab:alpha}, with $\alpha$ set to 0.2, we observed that the model achieved its optimal performance. An $\alpha$ value of 0 means the model directly utilizes the original dense CLIP features, while a setting of 1 implies complete reliance on reconstructed features. The results show that when relying entirely on reconstructed features, there is a significant decrease in model performance. This indicates that although reconstructed features help reduce the impact of noise, they also result in the loss of multimodal information in the CLIP features. While optimizing adapters through the method described in \Cref{adapter_opm_org} can mitigate the impact of noise to a certain extent, the supervision signal it relies on (that is, the original dense CLIP features) is inherently filled with a substantial amount of noise. In contrast, the rendered features are clearer, providing more effective supervision. Therefore, we attempt to use a self-cross training strategy as detailed in \Cref{adapter_opm_sct} to optimize the adapter, aiming to achieve better performance. In \Cref{tab:beta}, we explore the impact of different $\beta$ on model performance. Among them, when $\beta=0.3$, the model has the best performance and has a greater improvement compared to other hyperparameters.

\begin{table}
\caption{\textbf{Ablation studies} on 3D-OVS \cite{liu2023weakly}. We evaluate the adapter layer, the label volume (LV),  the self-cross-training (SCT) strategy, the low-rank transient query (TQ) attention, and the simplified text augmentation loss $\mathcal{L}_{aug}$. Our final method in the last row performs the best.}
\label{tab:ab_main}
\centering
\begin{tabular}{l|ll}
\toprule
Method                                         & \multicolumn{1}{c}{mIou} & \multicolumn{1}{c}{Acc.} \\ \hline
Base (only distill CLIP feature)               &      59.4                    &       74.5                   \\
+ adapter layer                                &       68.2                   &         83.4                 \\
+ adapter layer + SCT                 &                  73.9        &         89.7                 \\
+ adapter layer + SCT + TQ           &                   \cellcolor{third}78.6       &        \cellcolor{third}92.4                  \\
+ adapter layer + SCT + TQ + LV      &             \cellcolor{second}85.6           &         \cellcolor{second}96.0               \\
+ adapter layer + SCT + TQ + LV + $\mathcal{L}_{aug}$ &              \cellcolor{best}88.1     & \cellcolor{best}97.3      \\ \bottomrule
\end{tabular}
\end{table}

\begin{table}
\caption{\textbf{Ablations} on varying the residual ratio $\alpha$ for adapter layer.}
\label{tab:alpha}
\centering
\begin{tabular}{c|cccccc}
\toprule
Ratio $\alpha$ & 0 & 0.2 & 0.4 & 0.6 & 0.8 & 1.0 \\ \hline
mIou           &  81.8 & \cellcolor{best}88.1    &  87.6   &  87.3   &  87.2   &   71.8  \\
Acc.       &  94.3 & \cellcolor{best}97.3    &  96.2   &   96.0  &   95.9  &  90.6   \\ \bottomrule
\end{tabular}
\end{table}
\begin{table}
\caption{\textbf{Ablations} on varying the ratio $\beta$ for $\mathcal{L}_r$.}
\label{tab:beta}
\centering
\begin{tabular}{c|ccccccc}
\toprule
Ratio $\beta$ & 0 & 0.1 & 0.3 & 0.5 & 0.7 & 0.9 & 1.0\\ \hline
mIou           & 87.2  &  87.7   &  \cellcolor{best}88.1   &  87.8   &  86.5   &  85.9  & 85.8 \\
Acc.       & 95.9  &  96.8   &  \cellcolor{best}97.3   &  96.3   &   95.7  &  95.1   & 95.0\\ \bottomrule
\end{tabular}
\end{table}

\subsubsection{Investigation of Low-Rank Transient Query Attention} To investigate the complexity and performance of our low-rank transient query attention, we compared vanilla self-attention and without any attention block. As shown in the fourth column of \Cref{fig:ablation_main}, our low-rank transient query attention can enhance the information details of object edges, as we discussed in \Cref{sec:3}. However, for categories with similar colors, this approach may lead to additional misclassifications. For example, the black edges of graphics cards and black shoes are misidentified as the same category. The results are shown in \Cref{tab:tqa}, where D represents the dimensionality of the attention projection. By increasing the dimension of the hidden layer from 16 to 64, we found that excessively small dimensions significantly reduce model performance. The reason is that lower dimensions impair the capability to express semantic information of features. However, increasing the dimension also leads to a decline in performance, as excessively high dimensions introduce unnecessary redundancy. The optimal bottleneck dimension is 32, which can retain sufficient semantics without redundancy. The complexity of these two types of attention refers to \Cref{sec:3}. From the results, under the condition of maintaining the same dimensions, the traditional vanilla self-attention model has 18.18 times the floating-point operations (FLOPs) compared to our proposed low-rank transient query attention model, and 12 times our model in training time. Although our model has only 0.01MB more parameters than the vanilla self-attention, our mIoU score is improved by 0.02\%, and the accuracy score is improved by 0.12\%. Furthermore, compared to the baseline model without any integrated attention block, our model only adds 1 minute to the training time, while significantly improving the mIoU score and accuracy.

\begin{table}
\caption{\textbf{Ablations} on varying the low-rank transient query (TQ) attention, and Vanilla is the Vanilla self-attention.}
\label{tab:tqa}
\centering
\resizebox{\linewidth}{!}{
\begin{tabular}{c|ccccc}
\toprule
Method                   & mIou & Acc. & FLOPs $\downarrow$ & \# Params $\downarrow$ & Train Time $\downarrow$  \\ \hline
W/o attention &   85.31   &    95.41      &  -    & -  & 10 min\\
Vanilla (D=32) &   \cellcolor{second}87.98   &    \cellcolor{second}97.13      &  26.91G    & 0.25M  & 2.2 hrs\\
TQ (D=16)               &   37.18   &   83.58       &  0.69G    & 0.13M & 11 min\\
TQ (D=32)               &   \cellcolor{best}88.10   &    \cellcolor{best}97.25      &  1.48G   & 0.26M & 11 min\\
TQ (D=48)               &   \cellcolor{third}86.33   &    95.27      &  2.33G    & 0.39M & 11 min\\
TQ (D=64)               &   86.32  &    \cellcolor{third}95.55      &  3.28G    & 0.53M & 11 min\\ 
\bottomrule
\end{tabular}
}
\end{table}

\begin{figure}
    \centering
    \includegraphics[width=\linewidth]{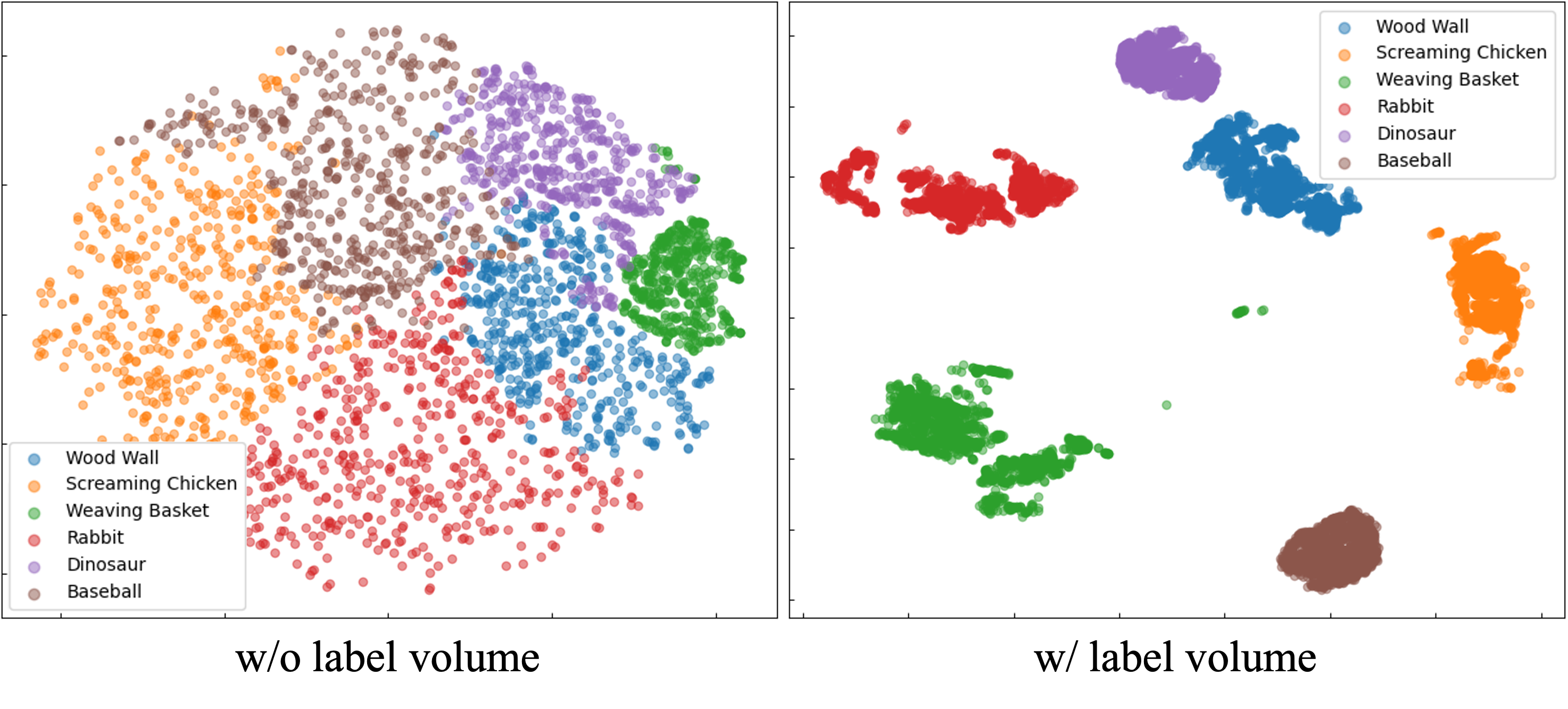}
    \caption{\textbf{Ablation} on the label volume. 3D points with similar feature expressions are gathered into the same category in the feature space, thereby enhancing the classification effect based on light consistency.}
    \label{fig:ablation_label}
\end{figure}

\begin{table}
\caption{\textbf{Ablations} on varying the ratio $\gamma$ for $\mathcal{L}_{CE}$.}
\label{tab:gamma}
\centering
\begin{tabular}{c|ccccc}
\toprule
Ratio $\gamma$  & 0.1 & 0.3 & 0.5 & 0.7 & 0.9 \\ \hline
mIou             &  84.1   &  87.3   &  \cellcolor{best}88.1   &  86.5   &  81.6  \\
Acc.         &  94.4   &   95.9  &  \cellcolor{best}97.3   &   95.7  &   93.1  \\ \bottomrule
\end{tabular}
\end{table}
\subsubsection{Investigation of Label Volume}
We utilized label volumes to transform the segmentation task into a classification problem, a change that is crucial for enhancing model performance. As shown in the fifth column of \Cref{fig:ablation_main}, the introduction of label volumes significantly optimized the segmentation results. In \Cref{fig:ablation_label}, we found that under the influence of label volumes, features of different categories exhibited a more compact and distinct distribution. This indicates that label volumes effectively improved the model's ability to discriminate the similarity between features of the same category, thereby making the feature representation of the same category more unified. This enhanced unity has led to a significant improvement in model performance. Furthermore, we delved into the specific impact of label volume at different weights on model performance. We observed that the model achieved optimal performance when $\gamma$ is set to 0.5. This can be attributed to the fact that label volume is achieved through the joint optimization of the original CLIP features and the rendered features. Overreliance on the original CLIP features may introduce excessive noise; conversely, overly relying on the rendered features may weaken the model's multimodal capability, thereby affecting its segmentation efficacy.

\begin{figure}
    \centering
    \includegraphics[width=\linewidth]{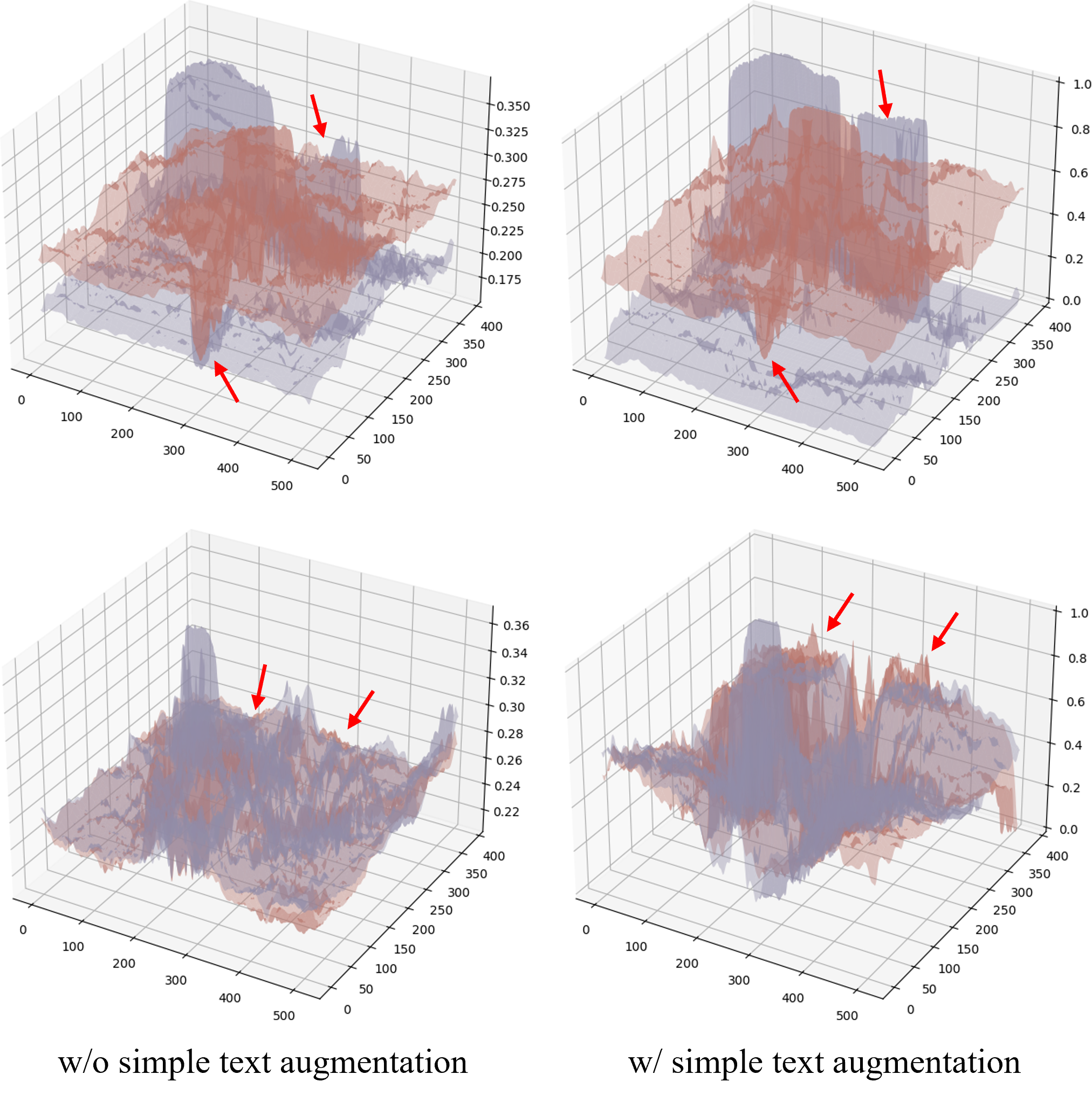}
    \caption{\textbf{Ablation} on simplified text augmentation. The x-axis and y-axis represent the pixel position of the image, while the z-axis indicates the degree of similarity between the text and the image. \textit{w/o simplified text augmentation}, there is notable ambiguity within the CLIP features indicated by the red arrows. Conversely, \textit{w/ simplified text augmentation}, the ambiguity in the CLIP features in these areas is reduced.}
    \label{fig:ablation_text}
\end{figure}

\subsubsection{Investigation of Simplified Text Augmentation Strategy}
As shown in the last column of \Cref{fig:ablation_main}, after adopting a simplified text augmentation strategy, the errors inside the graphics cards are significantly reduced, and the segmentation effect on the edges of the graphics cards is also noticeably improved. This is because the text augmentation strategy enhances the distinction between categories through repeated text enhancements and, after normalization, effectively alleviates the ambiguity problem caused by the high similarity between categories. The results of \Cref{fig:ablation_text} further confirm the effectiveness of our method: \textit{w/o simplified text augmentation strategy}, the locations marked with red arrows had extremely high category similarity, making it difficult to distinguish. However, \textit{w/ simplified text augmentation strategy}, the categories within the areas indicated by the red arrows are identified and differentiated.

\subsection{Limitations}
While our approach demonstrates notable enhancements in terms of speed and storage resource utilization when compared to prior methodologies, it is largely dependent on the predictive capabilities of dense pixel-level CLIP characteristics. The performance of our model is significantly influenced by the quality of these dense features, which frequently exhibit noise. As evidenced by our qualitative study, despite implementing many strategies to mitigate noise, our findings remain less refined and exhibit a certain level of noise in comparison to previously employed multi-scale features. This limitation becomes particularly evident when dealing with small objects or objects of similar colors. This suggests that while our approaches have made strides in noise reduction, the precision in identifying and segmenting objects under these challenging conditions requires further enhancement. Additionally,  the reason for the need to modify CLIP's encoder is due to the requirement of predicting dense pixel-level CLIP characteristics without any fine-tuning. While this technique retains its multi-modal capabilities, it is important to acknowledge that these adjustments will obviously have an impact on the performance of the model. Furthermore, our approach is limited in its ability to enhance model performance due to its inability to fine-tune the encoder utilizing extensive data sets, such as the 2D segmentation task. 

\section{Conclusion}
In this work, we propose Laser, an efficient language-guided segmentation of 3D radiation fields. Compared with previous research, we optimized the workflow and used dense pixel-level CLIP feature distillation to guide the radiation field to have text-guided segmentation capabilities. Specifically, we first combine adapter and self-cross-training strategies to effectively mitigate the noise introduced by dense pixel-level CLIP features. Next, we introduced low-rank transient query attention, which not only effectively enhanced the edge clarity of the segmentation map, but also reduced the complexity of vanilla self-attention in 3D feature processing. Furthermore, we enhance the consistency across different views by transforming the segmentation problem into a classification problem. Finally, we adopt a simplified text enhancement strategy to alleviate the ambiguity of CLIP correlations. Extensive experiments demonstrate that our method outperforms current state-of-the-art technologies in both speed and performance.

Our method is based on TensoRF \cite{chen2022tensorf} implementation and is limited in rendering speed, but a potential solution is to use state-of-the-art 3D Gaussian splatting technology \cite{kerbl3Dgaussians}. Furthermore, as we discussed in the limitation analysis, distilling directly from dense pixel-level CLIP features with noise will inevitably affect model performance. Therefore, a focus of our future research is to explore how to denoise these features more effectively. On the other hand, we plan to explore more general graph-based multimodal methods to more comprehensively capture the global relationship between different modalities. Currently, our method mainly focuses on the mining of local modal relationships but lacks the overall modal association. By introducing a more general graph-based method \cite{marcu2023self}, we hope to make up for this deficiency and thus improve the performance of the model in multimodal tasks.

\section*{Acknowledgments}
This work is supported by the UK Medical Research Council (MRC) Innovation Fellowship under Grant MR/S003916/2, International Exchanges 2022 IEC$\backslash$NSFC$\backslash$223523.

\bibliographystyle{IEEEtran}
\bibliography{main}

% \newpage

\end{document}